\documentclass{article} 
\usepackage{paper,times}
\iclrfinalcopy 

\usepackage{amsmath,amsfonts,bm}

\def\eqref#1{equation~\ref{#1}}

\def\1{\bm{1}}

\DeclareMathAlphabet{\mathsfit}{\encodingdefault}{\sfdefault}{m}{sl}
\SetMathAlphabet{\mathsfit}{bold}{\encodingdefault}{\sfdefault}{bx}{n}

\usepackage{changepage}
\usepackage{hyperref}
\usepackage{url}
\usepackage{graphicx}
\usepackage{enumitem}
\usepackage{booktabs}
\usepackage{algorithm}
\usepackage{algpseudocode}
\usepackage{multicol}
\usepackage{subfig}
\usepackage{amsmath, amsthm, amsfonts}
\usepackage{booktabs}
\usepackage{multirow}

\usepackage{lipsum} 
\usepackage{makecell}

\usepackage{xcolor}
\usepackage{xspace}
\usepackage{tabularx} 
\usepackage{wrapfig}

\begin{document}

\title{FT-Shield: A Watermark Against Unauthorized Fine-tuning in Text-to-Image Diffusion Models }


\author{
\textbf{~~~~ ~~ ~~ Yingqian Cui$^{1}$ ~~ Jie Ren$^{1}$ ~~ Yuping Lin$^{1}$ ~~ Han Xu$^{1}$ ~~ Pengfei He$^{1}$ ~~ Yue Xing$^{2}$ ~~} \\
\textbf{ ~~ ~~ ~~ ~~ ~~ ~~ ~~ ~~ ~~ Lingjuan Lyu$^{3}$ ~~ Wenqi Fan$^{4}$ ~~ Hui Liu$^{1}$ ~~ Jiliang Tang$^{1}$} \\
 ~~ ~~ ~~  ~~ ~~ $^{1}$Department of Computer Science \& Engineering, Michigan State University\\
 ~~ ~~ ~~ ~~  ~~ ~~ ~~ ~~$^{2}$Department of Statistics and Probability, Michigan State University\\
 ~~ ~~ ~~ ~~ ~~ ~~ ~~  ~~ ~~ ~~ ~~$^{3}$Sony AI ~~ ~~ $^{4}$The Hong Kong Polytechnic University \\
 ~~  ~~ \texttt{\small\{cuiyingq, renjie3, linyupin, xuhan1, hepengf1, xingyue1, liuhui7,} \\
 ~~  ~~ \small{\texttt{tangjili\}@msu.edu}} ~~\small{\texttt{wenqi.fan@polyu.edu.hk}} ~~
\small{\texttt{lingjuanlvsmile@gmail.com}}
}

\maketitle

%

\newcommand{\fix}{\marginpar{FIX}}
\newcommand{\new}{\marginpar{NEW}}

\newcommand{\han}[1]{\textcolor{violet}{Han: #1}}
\newcommand{\jie}[1]{\textcolor{blue}{jie: #1}}
\newcommand{\jt}[1]{\textcolor{red}{JT: #1}}
\newcommand{\yue}[1]{{\color{purple}#1}}


\begin{abstract}
Text-to-image generative models, especially those based on latent diffusion models (LDMs), have demonstrated outstanding ability in generating high-quality and high-resolution images from textual prompts. With this advancement, various fine-tuning methods have been developed to personalize text-to-image models for specific applications such as artistic style adaptation and human face transfer. However, such advancements have raised copyright concerns, especially when the data are used for personalization without authorization. For example, a malicious user can employ fine-tuning techniques to replicate the style of an artist without consent. In light of this concern, we propose FT-Shield, a watermarking solution tailored for the fine-tuning of text-to-image diffusion models. FT-Shield addresses copyright protection challenges by designing new watermark generation and detection strategies. In particular, it introduces an innovative algorithm for watermark generation. It ensures the seamless transfer of watermarks from training images to generated outputs, facilitating the identification of copyrighted material use. To tackle the variability in fine-tuning methods and their impact on watermark detection, FT-Shield integrates a Mixture of Experts (MoE) approach for watermark detection. Comprehensive experiments validate the effectiveness of our proposed FT-Shield.
\end{abstract}

\section{Introduction}

\label{intro}

Generative models, particularly Generative Diffusion Models (GDMs) \citep{ho2020denoising,song2020score,ho2022classifier,song2020denoising}, have witnessed significant progress in generating high-quality images from random noise. Recently, text-to-image 
generative models leveraging latent diffusion \citep{rombach2022high} have showcased remarkable proficiency in producing specific, detailed images from human language descriptions.
Based on this advancement, fine-tuning techniques such as DreamBooth \citep{ruiz2023dreambooth} and Textual Inversion \citep{gal2022image} have been developed. These methods enable the personalization of text-to-image diffusion models, allowing them to adapt to distinct artistic styles or specific subjects. For instance, with a few paintings from an artist, a model can be fine-tuned to adapt to the artistic style of the artist and create paintings which mimic the style.
However, the proliferation of these methods has sparked significant concerns 
about the potential misuse of these techniques for unauthorized style imitation or the creation of deceptive human facial images. Such actions can potentially violate creators' rights and compromise intellectual property (IP) and privacy integrity~\citep{chen2023pathway,wang2024did,wen2023detecting}.


Watermarking has emerged as a popular technique for protecting data's IP against various forms of infringement~\citep{cox2002digital,podilchuk2001digital,navas2008dwt,zhu2018hidden}. It works by injecting imperceptible signals or patterns into images which can later be identified by a watermark detector. This enables the tracking of unauthorized copies and facilitates the assertion of copyright infringement~\citep{cox2002digital,podilchuk2001digital,navas2008dwt,zhu2018hidden, ren2024copyright}. Compared to other protection methods such as encryption~\citep{forouzan2007cryptography}, watermarking enjoys several advantages such as its stealthy nature, resilience to manipulation, and the ability to trace and manage digital assets effectively. Given its advantages, the watermarking technique also has the potential for IP protection for text-to-image model fine-tuning. When the watermarked images are used for fine-tuning text-to-image models, the resulting generated images are expected to inherit the watermark, acting as an indelible signature. By employing the detector, the presence of the watermark in the generated content can be identified, providing evidence of IP infringement.

\begin{figure}
  \centering

\includegraphics[width=0.95\textwidth]{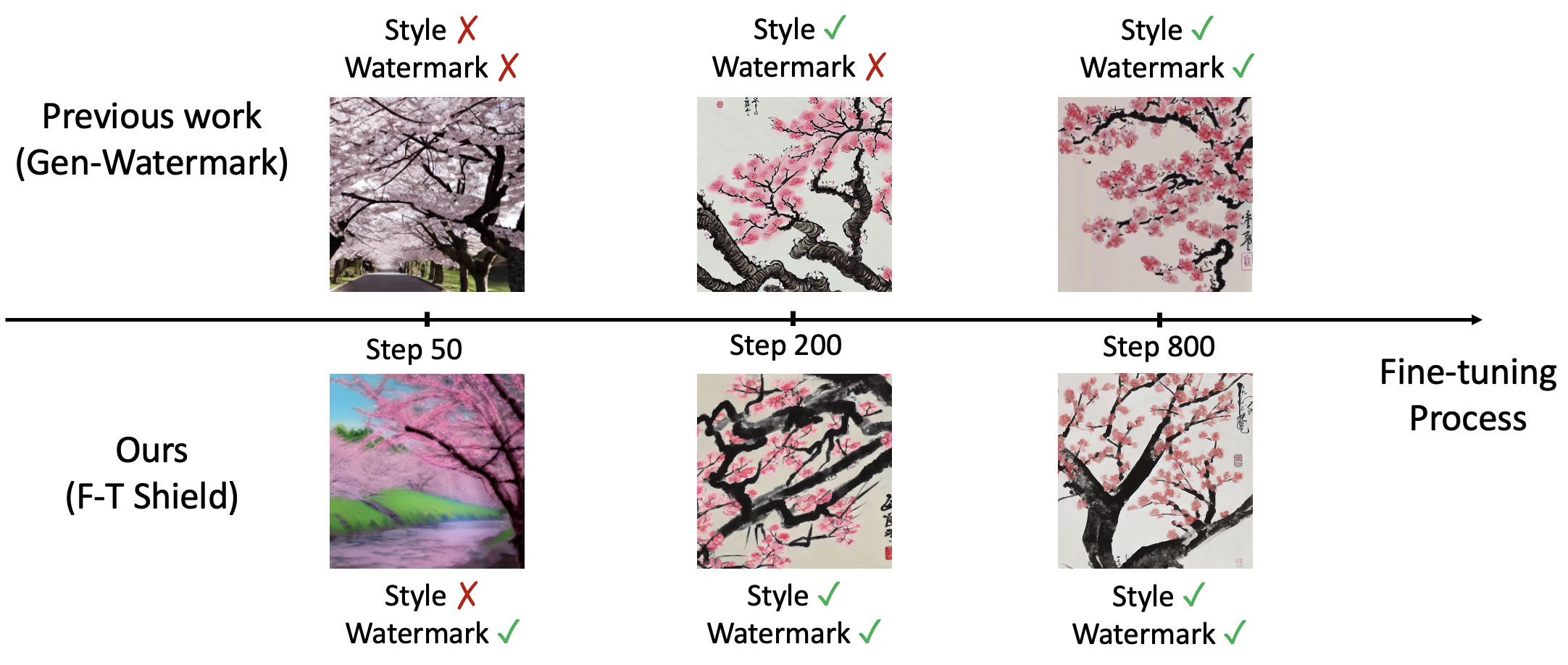} 

  \caption{\small{An illustration of generated images from fine-tuned text-to-image models w.r.t. fine-tuning steps. While previous work~\citep{ma2023generative} requires extensive fine-tuning to ensure that the watermark is learned, our method enables the watermark to be learned in the early stages of fine-tuning. Prompt used for the generation: Cherry blossoms in full bloom. Targeted style: the style of artist Beihong Xu.  }}

    \label{fig:intro}
\end{figure}

However, we face tremendous challenges in developing watermarking techniques to prevent unauthorized fine-tuning. First, according to prior theoretical findings~\citep{allen2022feature}, the fine-tuning process of neural networks involves a sequential pattern in feature learning: certain features can be learned earlier while others are acquired later. 
This indicates that designing what features to embed within the watermark is crucial. If the watermark's pattern cannot be assimilated by the model prior to style-related features, the infringer can easily circumvent the watermarks by reducing the fine-tuning steps, allowing the model to adopt the style without acquiring the watermark. Although there are recent watermarking methods proposed for images' IP protection against text-to-image model fine-tuning, 
this challenge has not been solved. As shown in Figure~\ref{fig:intro}, as the fine-tuning steps increase, the watermark proposed by \cite{ma2023generative} is acquired by the generative model later than the targeted style, indicating an ineffective protection provided by the watermark. One potential reason is that the watermark-generating procedure predominantly adheres to traditional watermark strategies, which are intended to trace the source of an image rather than protecting its IP in the context of diffusion models fine-tuning. Consequently, there's no assurance that the watermark's features will be learned by the model before other features.

Second, as indicated by~\cite{ma2023generative}, due to the distribution shift between the fine-tuning data of generative models and the resulting generated images, it is crucial to incorporate images generated by fine-tuned models to develop the watermark detector. Since there are numerous fine-tuning methods introduced from different perspectives, a detector for one method might lose its effectiveness for others. This issue is highlighted by~\cite{ma2023generative} that a large drop in watermark detection performance is observed when applying a detector tailored to one fine-tuning method to others. Meanwhile, in reality, a data protector may not know which fine-tuning method was used by the infringer, thus it is desired to design a watermark detection strategy that is effective for various fine-tuning methods.


To tackle the aforementioned challenges, we propose a novel watermarking framework, \textbf{FT-Shield}, tailored for data's copyright protection against the \textbf{F}ine-\textbf{T}uning of text-to-image diffusion models. In particular, we introduce a training objective incorporating the fine-tuning loss of diffusion models for watermark generation. By minimizing the objective, we ensure that the optimized watermark pattern can be quickly learned by diffusion model at the very early stage of fine-tuning. As shown in Figure~\ref{fig:intro}, even when the style has not been adopted, our watermark has already been learned by the diffusion model. Furthermore, to obtain a watermark detector for various fine-tuning methods, we introduce a detection framework based on  Mixture of Experts~\citep{jacobs1991adaptive}. The effectiveness of FT-Shield is verified through experiments across various fine-tuning approaches including DreamBooth~\citep{ruiz2023dreambooth}, Textual Inversion~\citep{gal2022image}, Text-to-Image Fine-Tuning~\citep{von-platen-etal-2022-diffusers} and LoRA~\citep{hu2021lora}, applied to both style transfer and object transfer tasks across multiple datasets.

\section{Related work}
\subsection{Text-to-image diffusion model and their fine-tuning methods}

Diffusion models \citep{ho2020denoising,song2020score,ho2022classifier,song2020denoising,dhariwal2021diffusion} have recently achieved remarkable advancements in the realm of image synthesis, notably after the introduction of the Denoising Diffusion Probabilistic Model (DDPM) by \cite{ho2020denoising}. Building upon the DDPM framework, \cite{rombach2022high} presented the Latent Diffusion Model (LDM). Unlike conventional models, LDM conducts the diffusion process within a latent space derived from a pre-trained autoencoder, and generates hidden vectors by diffusion process instead of directly generating the image in pixel space. This strategy enables the diffusion model to leverage the robust semantic features and visual patterns imbibed by the encoder. Consequently, LDM has set new standards in both high-resolution image synthesis and text-to-image generation. Building upon text-to-image diffusion models, multiple fine-tuning techniques~\citep{gal2022image, ruiz2023dreambooth,hu2021lora} have been developed. These methods enable the personalization of text-to-image models, allowing them to adapt to distinct artistic styles or specific subjects. Specifically, DreamBooth \citep{gal2022image} works by fine-tuning the denoising network of the diffusion model to make the model associate a less frequently used word-embedding with a specific subject. Textual Inversion \citep{ruiz2023dreambooth} tries to add a new token which is bound with the new concept to the text-embedding of the model. LoRA \citep{hu2021lora} adds pairs of rank-decomposition matrices to the existing weights of the denoise network and only trains the newly added weights in fine-tuning.




\subsection{Image protection methods}


To protect images' IP from unauthorized learning by text-to-image models, in literature, two predominant methods are employed: (1) Adversarial methods which design perturbations in the data to prevent any model learning from the data; and (2) Watermarking techniques which introduce imperceptible signals to the image to enable protectors to detect infringement. 

\textbf{Adversarial methods.}
Adversarial methods protect data's IP by applying the idea of evasion attacks. They treat the unauthorized generative models as the target of attack, and develop adversarial examples to disrupt the learning process of unauthorized fine-tuning.
GLAZE~\citep{shan2023glaze} is the first adversarial method which focuses on attacking the features extracted by the encoder in Stable Diffusion to prevent the learning of image styles.
The work of~\cite{van2023anti} and \cite{liang2023adversarial} introduces methods to generate adversarial examples to evade the infringement from DreamBooth~\citep{ruiz2023dreambooth} and Textual Inversion~\citep{gal2022image}, respectively.
Additionally,~\cite{salman2023raising} proposed to alter the pictures to protect them from image editing applications by Stable Diffusion in case the pictures are used to generate images with illegal or abnormal scenarios.
Although these methods provide effective protection against infringement, they can inadvertently disrupt authorized uses (such as for academic research purposes) of the safeguarded images. This indicates the necessity of developing watermarking approaches, which allow the IP to be used for proper reasons while also acting as a way to collect proof against improper uses. 

\textbf{Watermarking methods.}
Watermark has also been considered to protect the IP of images against unauthorized usage during the fine-tuning of text-to-image models. \cite{wang2024diagnosis} proposed to apply an existing backdoor method~\citep{nguyen2021wanet} to embed unique signatures into the protected images. It aims to inject extra memorization into the text-to-image models fine-tuned on the protected dataset so that unauthorized data usage can be detected by checking whether the extra memorization exists in the suspected model. However, one limitation is the assumption that the suspicious model is readily accessible to the data protector. This might not be realistic, as malicious entities can hide the fine-tuned model and only disclose a handful of generated images. Another work by~\cite{ma2023generative} introduces a method that simultaneously trains a watermark generator and detector. The detector is then further fine-tuned with the images generated by the fine-tuned model. However, as discussed in Section~\ref{intro}, the issue of this technique is that it provides no guarantee that their watermark can be learned by the model earlier than the image's style. The shortcomings of current methods highlight the need for a more robust and effective watermarking design.

\section{Method}

In this section, we first define the problem and introduce the key notations. 
Then we elaborate on some prior theoretical insights to present the challenge in watermark generation and introduce details of the watermark generation process employed by FT-Shield. Lastly, we introduce the MoE framework to detect watermarks on images generated by various fine-tuning methods.
\begin{figure}[tb]
  \centering
\includegraphics[width=0.95\textwidth]{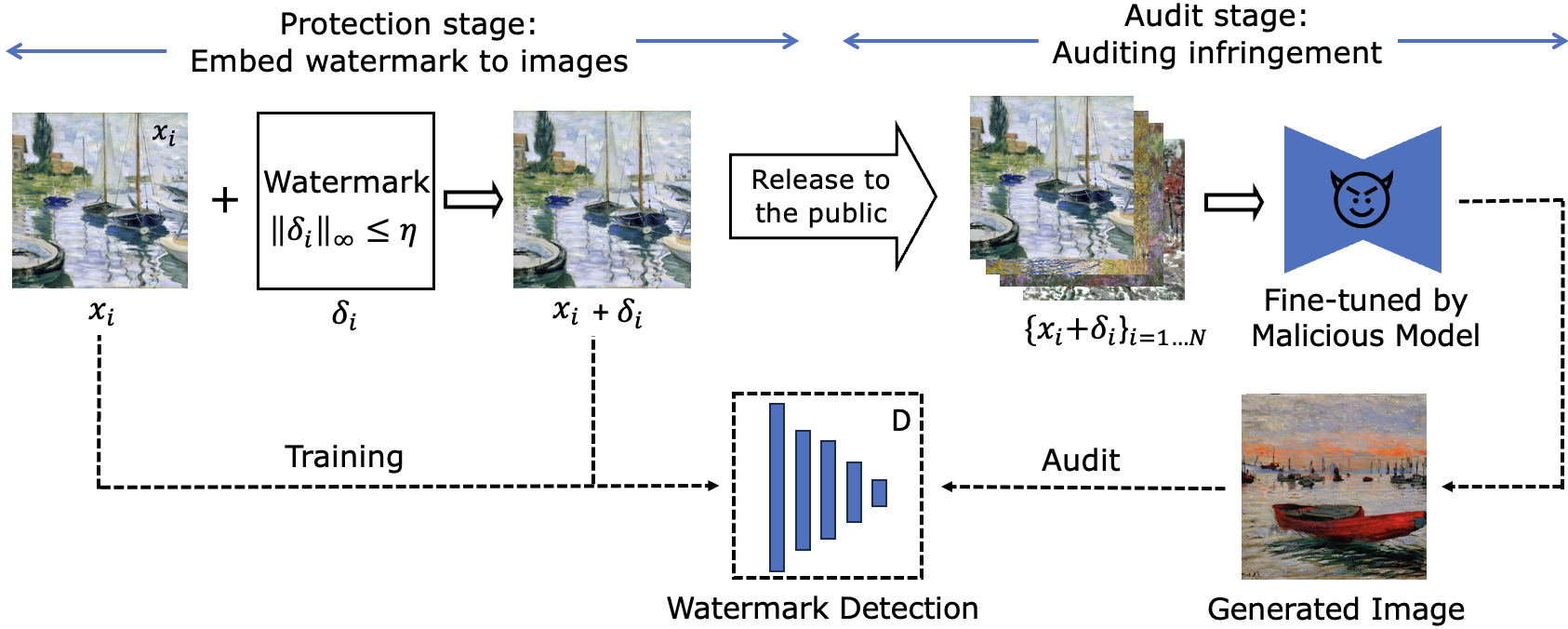} 
  \caption{\small An overview of the two-stage watermarking protection process}
\label{fig:flow}
\end{figure}
\subsection{Problem formulation}
In the scenario of copyright infringement and protection considered in this work, there are two roles: 
(1) a \textbf{data protector} that possesses the data copyright, utilizes watermarking techniques before the data is released, and tries to detect if a suspected image is generated by a model fine-tuned on the protected images, and
(2) a \textbf{data offender} that uses the protected data for text-to-image model fine-tuning without permission from the data protector. The data offenders have complete control over the fine-tuning and sampling processes of the text-to-image diffusion models, while the data protectors can only modify the data they own before their data is released and access images generated by the suspected model.

As shown in Figure~\ref{fig:flow}, the protection process consists of two stages: the protection stage and the audit stage.
In the \textbf{protection stage}, the data protector protects the images by adding imperceptible watermarks to the images. Specifically, given that the size of the protected dataset is $N$, the target is to generate sample-wise watermark $\delta_i$ for each protected image $x_i,\forall i=1...N$. Then these watermarks are embedded into the protected images $\hat{x}_i=x_i+\delta_i$. Correspondingly, the data protector develops a watermark detection approach, denoted by function $D_{w}(\cdot)$, to test whether there is a watermark on the suspected image.
To ensure that the watermarks will not lead to severe influence on image quality, we limit the budget of the watermark by constraining its 
$l_\infty$ norm ($\|\delta_i\|_\infty \leq  \eta$) to control the pixel-wise difference between the two images $x_i$ and $\hat{x}_i$. 
In the \textbf{audit stage}, if the protectors encounter suspected images potentially produced through unauthorized text-to-image models fine-tuning, they will apply the watermark detection process $D_{w}(\cdot)$ to ascertain whether these images have infringed upon their data rights.

\subsection{Watermark generation}
As mentioned in Section~\ref{intro}, to ensure robust IP protection, it is crucial for the watermarks to be learned earlier in the fine-tuning process. In this subsection, we first introduce the key challenge to achieve this goal by revising how neural networks learn features in pre-training and fine-tuning. Then we propose our watermark generation approach, explaining how it effectively mitigates this challenge.

\label{sec:gen_wm}
\textbf{A key challenge for watermark generation.} Based on the theoretical analysis on \cite{allen2022feature}, due to the random initialization, each hidden node in neural networks randomly captures some features at the beginning of training. During pre-training, instead of learning other features from the data, the network emphasizes and strengthens those features that were captured at initialization and are also present in the dataset. Meanwhile, it eliminates features that, despite being learned at initialization, do not find a match in the dataset. Essentially, a well-pre-trained model captures only the features that appear in the pre-training data.
An important implication of \cite{allen2022feature} is that, when learning from the fine-tuning data, the neural network can easily adapt the features that already exist in the pre-training data, but it is difficult to learn new features that only appear in the fine-tuning data. This suggests a challenge: if the watermark's features are new to the diffusion model, it is hard to ensure that the watermark can be easily learned by the model during fine-tuning.

\textbf{The proposed watermark generation approach}. To overcome the above challenge, we propose to simultaneously train the watermark and fine-tune the model as follows. Given $N$ samples to be protected, we construct a training objective for the watermark as:
\begin{align}
    \min_{\theta_1}\min_{\{\delta_i\}_{i\in[N]}} \sum_{i\in[N]} L_{LDM}(\theta_1,\theta_2,x_i+\delta_i,c) \text { s.t. } \|\delta_i\|_\infty \leq  \eta 
\label{eqn:obj}
\end{align}
where $\theta_1$ represents the parameters of the UNet~\citep{ronneberger2015unet}, which is the denoise model within the text-to-image model structure, $\theta_2$ denotes the parameters of the other part of the diffusion model, and $c$ is the prompt for the image generation. The function ${L}_{d m}$ indicates fine-tuning loss of the text-to-image diffusion model:
\begin{align}
L_{LDM}(\theta_1, \theta_2, x_0, c) = \mathbb{E}_{t, \epsilon \sim \mathcal{N}(0,I_d)}\left\|\epsilon - \epsilon_{\theta_1,\theta_2}(x_t, t, c)\right\|_2,\label{eqn:detect}
\end{align}
with $d$ as the dimension of the training images in the latent space, $x_0$ as the input image, $t$ as the timestep and $x_t$ as the input image with $t$ steps' noise in the diffusion process.
The above training objective aims to identify the best perturbation $\delta_i$ for each sample $x_i$ so that the loss of a diffusion model trained on these perturbed samples can be minimized. 
%

To explain the above design in Equation~\ref{eqn:obj}, since different features are learned differently in the fine-tuning stage, we use a minimization to find the most proper features for the watermark. The inner minimization of $\delta_i$ ensures that the watermark can be easily captured by the fine-tuned model, even when the fine-tuning does not involve many steps. Numerically, we solve this bi-level optimization problem by alternatively updating the perturbation and the parameters of the diffusion model. Details about the algorithm are provided in Appendix~\ref{append:alg}.

{\bf The design of fine-tuning loss and prompt.} 
There have been multiple fine-tuning methods for text-to-image diffusion models and each of them involves different fine-tuning loss and different parts of the model to be updated. 
Therefore, it is crucial to ensure that our watermark remains effective across different types of fine-tuning. This requires a careful selection of the specific $L_{LDM}$ in Equation~\ref{eqn:obj} and the caption $c$ used in watermark training. In terms of the particular $L_{LDM}$ in Equation~\ref{eqn:obj}, to maintain simplicity and coherence, we focused on the most fundamental Text-to-Image Fine-Tuning Method~\citep{von-platen-etal-2022-diffusers}. This method aligns with the original training objectives of the text-to-image model and involves updates only to the UNet. Our experiments in Section~\ref{sec:results} demonstrate that the watermarks can be assimilated well by different fine-tuning methods, even for those which did not modify the UNet such as Textual Inversion~\citep{ruiz2023dreambooth}.
For the caption $c$,  we employ a simplistic and consistent caption format for every image in the dataset in watermark generation. This consistency ensures robustness in varying conditions. Specifically, for the images associated with style transfer tasks, the caption is `\textit{A painting by *}' for art datasets, with * denoting the artist's name, and `\textit{A Pokemon character}' for the Pokemon dataset. For images used for object transfer, each is labeled by `\textit{a photo of *}', where * indicates the object's category, such as `toy', or `person'.




\subsection{Mixture of Watermark Detectors}
\label{mtd:moe} 
\begin{figure}[tb]

  \centering

\includegraphics[width=0.71\textwidth]{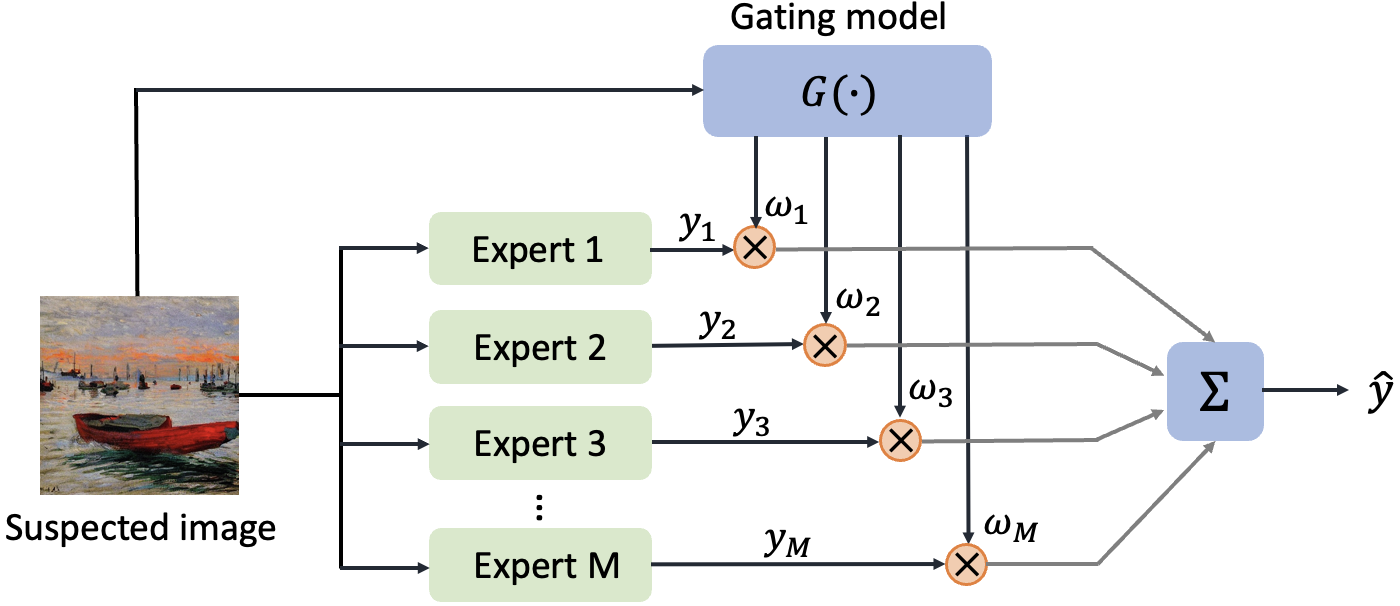} 

  \caption{\small An illustration of mixture of watermark detectors.}
\label{fig:flow_moe}
\end{figure}

\label{detector}
After the watermarks are generated, we need to detect if a suspected image contains a watermark or not. As mentioned in Section~\ref{intro}, a challenge of the watermark detection lies in the problem that the watermark detector tailored to one fine-tuning method cannot transfer well to other methods. This is because different fine-tuning methods modify various parts of the model, resulting in distinct impacts on the generation process and feature updates. To address this issue, we propose a strategy based on the Mixture of Experts (MoE)~\citep{jacobs1991adaptive}. 

\textbf{The general workflow.} An overview of the watermark detection process using MoE is presented in Figure~\ref{fig:flow_moe}. The MoE framework comprises two main elements: the Gating Model and multiple expert models.
In the watermark detection phase, the Gating Model is applied to the suspected image to estimate the likelihood that the image was produced by each potential fine-tuning method. Meanwhile, multiple experts of watermark detector, each customized for a distinct fine-tuning method, are employed. The final prediction on the presence of a watermark is then derived by taking a weighted average of these expert assessments, with the weights determined by the Gating Model's probabilities. Formally, the MOE detection framework can be formulated as:

\begin{align*}
    D_{w}(x) = \sum_{i=1}^{M}\text{softmax}_i(G(x))\cdot E_i(x),
\end{align*}

where $x$ refers to the image to evaluate, $M$ refers to the total number of expert models, $G(\cdot)$ denotes the Gating Model, $E_i(\cdot)$ represents the i-th expert model, and $\text{softmax}_i(\cdot)$ is the i-th output of the softmax function. 

\textbf{Two-stage training.} Instead of training the experts and Gating Model simultaneously, we employ a two-stage training strategy for watermark detection. Specifically, we first train each expert individually with the binary cross-entropy loss. Following the work of~\cite{ma2023generative}, we enhance the training dataset for each expert with images generated by fine-tuned text-to-image models. This process unfolds as follows: we first fine-tune the text-to-image model using a particular fine-tuning method with both clean and watermarked datasets, yielding two separately fine-tuned models. Subsequently, these models are employed to generate two distinct sets of data. Data generated from the model fine-tuned with clean data are incorporated into the original clean dataset, and the images generated from the model fine-tuned with watermarked images are utilized to augment the watermarked dataset for the training of the detector. After the experts are developed, we then train the Gating Model with a cross-entropy loss function, leveraging a dataset organized into various classes, where each class contains images generated by a specific fine-tuning method.

 This two-stage training strategy has several advantages. First, it provides good adaptability to new fine-tuning methods for text-to-image models. Updating the MoE system requires only the training of the new experts and the Gating Model. This enables the straightforward integration of previously trained experts without necessitating their retraining. Second, it effectively prevents the potential ``collapse problem''~\citep{shazeer2017outrageously} of MoE, which refers to the situation that the prediction relies on only the output of a single expert. Training the experts separately and initially helps to prevent this issue, promoting a more equitable and effective engagement of all experts. Third, it has a good memory efficiency. Since each expert is trained and inferred independently, it is unnecessary to load all experts into memory at once. 

\vspace{-0.01cm}
\section{Experiment}
\vspace{-0.01cm}
In this section, we evaluate the effectiveness of FT-Shield across various fine-tuning methods, subject transfer tasks and different datasets. We first introduce our experimental setups in Section~\ref{sec:setup}. 
In Section~\ref{sec:results} and Section~\ref{sec:moe}, we evaluate and analyze the detection performance of FT-Shield. Then we assess the impact of FT-Shield on image quality in Section~\ref{exp:quality}.
We further investigate our approach in Section~\ref{sec:steps} and Section~\ref{sec:robust} for its performance under fewer fine-tuning steps and robustness against image corruptions. Due to space limitation, our ablation studies on FT-Shield's performance with a reduced watermark rate and without training the detector on images from fine-tuned models are detailed in Appendix~\ref{appd:ablation}.

\subsection{Experimental settings}
\label{sec:setup}

\textbf{Model, Task and Dataset.}  We use Stable Diffusion as the pre-trained text-to-image model.
The image size is $512\times512$. We mainly focus on two tasks: style transfer and object transfer. Within style transfer, we explore two subtasks: one centers on art, utilizing 10 diverse datasets from WikiArt, each containing 20 to 40 images; the other focuses on a popular culture, employing the Pokémon BLIP captions dataset~\citep{pinkney2022pokemon}, which includes 833 images. 
The object transfer task involves two datasets of lifeless objects from~\cite{ruiz2023dreambooth} and three datasets of individual human faces from CelebA~\citep{liu2015faceattributes}, each comprising five images. We adopt different fine-tuning methods for different tasks. For style transfer, the methods include DreamBooth~\citep{ruiz2023dreambooth}, Textual-inversion~\citep{gal2022image}, Text-to-Image Fine-Tuning~\citep{von-platen-etal-2022-diffusers}, and LoRA~\citep{hu2021lora}, while for object transfer, only DreamBooth and Textual-inversion are utilized because the performance of the other two methods is not satisfying. 

\textbf{Baselines and Scenarios.} Our baselines include  Gen-Watermark~\citep{ma2023generative} and DIAGNOSIS~\citep{wang2024diagnosis}, which are also watermarking methods for IP protection against the fine-tuning of text-to-image model. 
The two baselines involve different settings in watermark detection. Gen-Watermark~\citep{ma2023generative} requires images generated from fine-tuned models to develop watermark detectors, resulting in the requirement of different watermark detectors tailored to different fine-tuning methods. In contrast, DIAGNOSIS~\citep{wang2024diagnosis} utilizes a general watermark detector to evaluate images generated by different fine-tuning methods. It does not require the knowledge about which specific method is used to generate the suspected image.

To compare with the two baseline methods, we evaluate FT-Shield considering two scenarios: 1) the data protector knows the specific fine-tuning method used by the offender; and 2) the protector is unaware of the fine-tuning method. In the first scenario, the protector can directly apply the watermark detector tailored to that fine-tuning method. We denote our method in this setting as \textbf{FT-Shield-Specific} and compare it with Gen-Watermark~\citep{ma2023generative}. In the second case, we employ the watermark detection based on MoE, denoted it as \textbf{FT-Shield-MoE}, and compare it with DIAGNOSIS~\citep{wang2024diagnosis}.



\textbf{Implementation Details.}
For watermark generation, we consider watermark budgets of 4/255 and 2/255 for each dataset. The watermarks are trained with 5-step PGD~\citep{madry2017towards} with the step size to be 1/10 of the budget. For training the experts and Gating Model used in MoE, we adopt ResNet18~\citep{he2016deep} with the Adam optimizer~\citep{kingma2014adam}, taking the learning rate as 0.001 and the weight decay as 0.01. In the detection stage, we use 60 and 30 prompts for image generations in style transfer and object transfer tasks, respectively.
Details about the prompts and hyperparameters of the fine-tuning methods are in Appendix~\ref{appd:prompts} and~\ref{append:exp}.

\textbf{Evaluation Metrics.} 
We evaluate FT-Shield from two perspectives: 1) its detection performance on data generated by fine-tuned models and 2) its influence on the quality of the released protected images and the generated images. For \textit{Detection Performance}, we consider two metrics. First, the detection rate (or true positive rate, TPR) quantifies the proportion of instances where the detector accurately identifies images produced by models fine-tuned on watermarked images. 
Second, the false positive rate (FPR) indicates the rate of instances where the detector mistakenly flags images without watermarks as watermarked. For \textit{Image Quality}, we use FID~\citep{heusel2017gans} for evaluation. Specifically, we measure the visual discrepancies between the original and watermarked images to evaluate its influence on the released images' quality. We also calculate the FID between the images generated from models fine-tuned on clean images with those generated from models fine-tuned on watermarked images to measure the watermark's influence on the generated images. A lower FID indicates better invisibility. 


\begin{table}[tb]
    \centering
    \captionsetup{font=small}
    \caption{Detection performance of FT-Shield-Specific}

    \label{main:detection}
    \resizebox{0.8\textwidth}{!}{
    
    \begin{tabular}{ll|cc|cc|cc}
        \toprule
        \multicolumn{1}{c}{} & \multicolumn{1}{c|}{} & \multicolumn{2}{c|}{ours ($\eta=4/255$)} & \multicolumn{2}{c|}{ours ($\eta=2/255$)} & \multicolumn{2}{c}{\makecell{Gen-Watermark\\ \citep{ma2023generative}} } \\
        \midrule
        \multicolumn{1}{c}{} & \multicolumn{1}{c|}{} & TPR$\uparrow$ & FPR$\downarrow$ & TPR$\uparrow$ & FPR$\downarrow$ & TPR$\uparrow$ & FPR$\downarrow$ \\
        \midrule
        \multirow{4}{*}{     \makecell{Style\\ (Arts)} }&DreamBooth& \textbf{99.50\%} & \textbf{0.18\%} & 98.68\% &0.87\%   & 93.31\% & 3.81\%  \\
        & Textual Inversion & \textbf{96.12\%} & \textbf{3.03\%} & 93.55\% & 5.25\% & 78.75\% & 12.70\% \\
        & Text-to-image     & \textbf{98.77\%} &\textbf{1.28\%} & 96.77\% & 3.54\% & 75.41\% & 30.03\% \\
        & LoRA              & \textbf{97.65\%} &\textbf{2.67\%} & 93.37\% & 6.17\% & 67.28\% & 22.72\% \\
          \midrule
        \multirow{4}{*}{\makecell{Style\\ (Pokemon)}}&DreamBooth& \textbf{99.5\%} & 0.50\% & 98.67\% & \textbf{0.33\%}  & 95.50\% & 3.67\%  \\
        & Textual Inversion & {96.00\%} &\textbf{1.67\%} & 94.67\% & 4.33\% & \textbf{97.83\%} & 2.67\% \\
        &Text-to-image     & \textbf{100.00\%} &\textbf{0.00\%} & 99.83\% &  \textbf{0.00\%}   & 98.33\%&3.54\% \\
        &LoRA              & 99.67\% &\textbf{0.00\%} & \textbf{99.83\%} & \textbf{0.00\%} & 97.71\% & 3.54\%\\
        \midrule
        \multirow{2}{*}{Object}
        & DreamBooth        & \textbf{98.93\%} & 1.23\% & 97.60\% & \textbf{1.13\%} & 91.39\% & 3.50\% \\
        & Textual Inversion & \textbf{97.73\%} & \textbf{1.67\%} &97.23\% & 1.97\% & 88.22\% & 3.95\%  \\
        \bottomrule
    \end{tabular}
      }
  
{\small (``$\uparrow$'' means a higher value is better. ``$\downarrow$'' means a lower value is better.)}

\end{table}

\subsection{Effectiveness of FT-Shield-Specific}

\label{sec:results}
\label{exp:main}
In this experiment, we evaluate the performance of FT-Shield-Specific.
The average of the TPR and FPR across multiple datasets for different transfer tasks are demonstrated in Table~\ref{main:detection}.
According to the results in Table~\ref{main:detection}, our method is able to protect the images with the highest TPR and lowest FPR
among most of the fine-tuning methods in both style and object transfer tasks. With an $l_\infty$ budget of 4/255, the TPR nearly reaches 100\% across all fine-tuning methods, while the FPR is close to 0. Even constrained by a small budget (2/255), FT-Shield can still achieve a TPR consistently higher than $90\%$ and an FPR no higher than $7\%$.
In comparison, Gen-Watermark~\citep{ma2023generative} has decent performance in the Pokemon style transfer and object transfer tasks but fails to achieve good performance in the art style transfer task. This indicates that it cannot consistently provide reliable protection across different applications.

\textbf{Transferability of Tailored Watermark Detector.} We explore the transferability of tailored watermark detectors by assessing their performance on images generated by other fine-tuning methods.
The performance of FT-Shield-Specific and Gen-Watermark~\citep{ma2023generative} on style transfer tasks is shown in Table~\ref{exp:trans_object}. The results of object transfer tasks are provided in Appendix~\ref{appd:ablation}. From Table~\ref{exp:trans_object}, it can be observed that generally FT-Shield-Specific outperforms Gen-Watermark~\citep{ma2023generative} in terms of the transferability across different fine-tuning methods. Nonetheless, FT-Shield-Specific still experiences an obvious performance drop when the detectors tailored for one fine-tuning method are applied to images generated by other methods. This highlights the need for a method to enhance the adaptability of the detector to different fine-tuning methods. The experiments in Section~\ref{sec:moe} will demonstrate that FT-Shield-MoE can mitigate this problem.

\begin{table}[tb]
    \centering
    \captionsetup{font=small}
    \caption{\small Transferability of the tailored watermark detectors in Style Transfer (Arts). Each number indicates the classifier's detection accuracy (average of true positive rate and true negative rate) when trained on images generated by the column's fine-tuning method and applied to images from the row's fine-tuning method.}

    \resizebox{0.83\textwidth}{!}{
        \begin{tabular}{l|l|cccc}
            \toprule
          Method & & DreamBooth & Textual Inversion & Text-to-Image & LoRA \\
            \midrule
            \multirow{4}{*}{\makecell{Ours  (2/255)}} & DreamBooth        & 98.91\%&68.94\% & 79.29\% & 78.04\%  \\
            &Textual Inversion & 70.52\% & 94.15\% &57.42\%& 58.38\%\\
            &Text-to-Image     &87.27\%  &  57.02\%&96.62\%&91.15\% \\
           & LoRA              & 91.50\% &  63.79\% &93.19\%& 93.60\%
           \\
           \midrule
            \multirow{4}{*}{\makecell{Ours (4/255)}} & DreamBooth        & 99.66\% &87.02\% & 84.38\% & 83.50\%  \\
            &Textual Inversion & 88.27\% &  96.55\% &68.96\%& 72.48\%\\
            &Text-to-Image     &94.35\%  &  68.60\%&  98.75\% &93.73\% \\
           & LoRA              & 97.71\% &  73.21\% &97.54\%&97.49\%\\
           \midrule
            \multirow{4}{*}{\makecell{Gen-Watermark\\~\citep{ma2023generative}}} & DreamBooth        & 95.92\% & 64.01\% & 53.87\% &64.11\%  \\
            &Textual Inversion & 68.01\% & 83.03\% &60.23\%& 75.39\%\\
            &Text-to-Image     &57.10\%  &  63.94\%& 72.69\% &63.77\% \\
           & LoRA              & 70.75\% &  66.29\% &54.39\%&72.28\% \\
            \bottomrule
        \end{tabular}
        \label{exp:trans_object}
    }

\end{table}

\begin{figure}[t]
  \centering

  \includegraphics[width=0.95\textwidth]{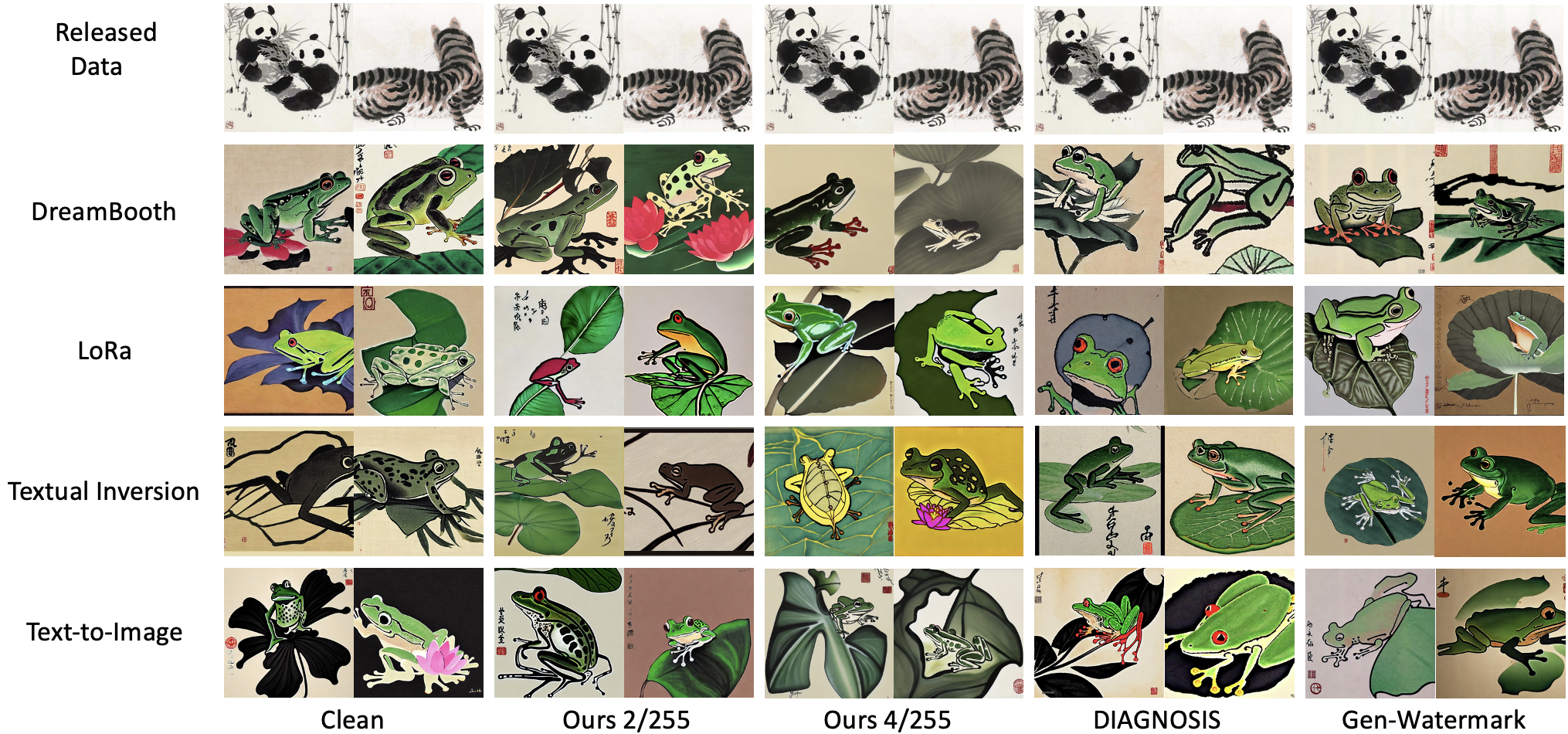}

  \captionsetup{font=small}
  \caption{\footnotesize{Examples of watermarked images (first line) and generated images (other lines) in the style of artist Beihong Xu. The prompt of generation: A frog on a lotus Leaf.}}
  
  \label{fig:examples} 

\end{figure}

\subsection{Effectiveness of FT-Shield-MoE}

\label{sec:moe}
\begin{table}[tb]
    \centering
    \captionsetup{font=small}
    \caption{Performance of FT-Shield-MoE}

    \label{exp:moe}
    \resizebox{0.75\textwidth}{!}{
    
    \begin{tabular}{ll|cc|cc|cc}
        \toprule
        \multicolumn{1}{c}{} & \multicolumn{1}{c|}{} & \multicolumn{2}{c|}{ours ($\eta=4/255$)} & \multicolumn{2}{c|}{ours ($\eta=2/255$)} &
        \multicolumn{2}{c}{\makecell{DIAGNOSIS\\ \citep{wang2024diagnosis}}} \\
        \midrule
        \multicolumn{1}{c}{} & \multicolumn{1}{c|}{} & TPR$\uparrow$ & FPR$\downarrow$ & TPR$\uparrow$ & FPR$\downarrow$ & TPR$\uparrow$ & FPR$\downarrow$ \\
        \midrule
        \multirow{4}{*}{\makecell{Style\\ (Arts)}}&DreamBooth& {\textbf{99.42\%}} & \textbf{0.83\%} & 97.79\% & 1.83\% &84.27\% & 4.18\% \\
        & Textual Inversion & {95.37\%} & 0.58\% & \textbf{95.67\%} & 3.04\% & 68.07\% & \textbf{0.25\%} \\
        & Text-to-image     & \textbf{99.04\%} &\textbf{1.79\%} & 96.83\% & 3.12\% &71.87\% & 5.30\% \\
        & LoRA              & \textbf{97.21\%} &\textbf{2.87\%} & 91.08\% &6.79\% &67.97\% & 9.78\%\\
          \midrule
        \multirow{4}{*}{\makecell{Style\\ (Pokemon)}}&DreamBooth& \textbf{99.00\%} & 2.17\% & 98.67\% & 2.50\% & 48.11\% & \textbf{0.67\%} \\
        & Textual Inversion & \textbf{92.33\%} & \textbf{1.67\%} & 91.17\% & 4.17\% & 57.05\% & 4.00\% \\
        & Text-to-image     & \textbf{99.83\%} &{1.17\%} & \textbf{99.83\%} &  \textbf{0.00\%}& 82.05\% & 6.67\% \\
        & LoRA              & \textbf{99.67\%} &\textbf{0.67\%} & \textbf{99.67\%} &\textbf{0.67\%}  & 83.56\% & 7.83\% \\
        \midrule
        \multirow{2}{*}{Object}
        & DreamBooth        & \textbf{99.33\%} & \textbf{1.08\%} & 98.08\% & {2.17\%} & 75.97\% & 1.20\% \\
        & Textual Inversion & {97.00\%} & {2.08\%} &\textbf{98.17\%} & \textbf{1.67\%} & 58.41\% & 22.20\% \\
        \bottomrule
    \end{tabular}
      }
      \end{table}

In this subsection, we demonstrate the effectiveness of FT-Shield-MoE.
The detection performance of FT-Shield-MoE when applied to images generated by different fine-tuning methods is shown in Table~\ref{exp:moe}. According to the results, FT-Shield-MoE demonstrates outstanding performance across various datasets. It consistently outperforms the baseline method DIAGNOSIS~\citep{wang2024diagnosis} across different fine-tuning methods and different transfer tasks. For most of the fine-tuning methods, the TPR is consistently higher than 90\% and the FPR is consistently lower than 5\%, which is comparable to the performance of FT-Shield-Specific under the scenario that fine-tuning method is known to the protector. 


\begin{table}[tb]

    \centering
    \captionsetup{font=small}
    \caption{\small FID $\downarrow$ between clean and watermark images in released and generated images}

    \label{main:fid}
    \resizebox{0.9\textwidth}{!}{
        \begin{tabular}{lllcccc}
        \toprule
        & & & \makecell{FT-Shield \\ ($\eta:4/255$)} & \makecell{FT-Shield \\ ($\eta:2/255$)} & \makecell{Gen-Watermark\\~\citep{ma2023generative} }  & \makecell{DIAGNOSIS\\~\citep{wang2024diagnosis}} \\
        \midrule
        \multirow{5}{*}{ \makecell{Style\\ (Arts)}} & \multicolumn{1}{c}{Released} & & 20.80 & \textbf{6.79} & 58.04 & 65.50 \\
        \cmidrule(lr){2-7}
        & \multirow{4}{*}{~~Generated~~} & DreamBooth & 62.25 & 46.96 & \textbf{46.42} & 49.77 \\
        &                     & Textual Inversion & 67.99 & 59.25 & \textbf{41.99} & 62.73 \\
        &                         & Text-to-image & 33.66 & \textbf{33.00} & 38.40 & 35.71 \\
        &                                  & LoRA & 32.76 & \textbf{29.99} & 30.12 & 33.30 \\
        \midrule
          \multirow{5}{*}{\makecell{Style\\ (Pokemon)}} & \multicolumn{1}{c}{Released} & & 27.93 & \textbf{10.63} & 57.90 & 21.03 \\
        \cmidrule(lr){2-7}
        & \multirow{4}{*}{~~Generated~~} & DreamBooth & 43.53 & 38.14 & \textbf{32.22} & 34.86 \\
        &                     & Textual Inversion & 96.98 & \textbf{45.76} & 67.24 & 67.21\\
        &                         & Text-to-image & 32.52 & \textbf{27.22} & 47.54 & 27.73 \\
        &                                  & LoRA & 39.53 & \textbf{33.52} & 49.25 & 38.82 \\
        \midrule
        \multirow{3}{*}{Object} & \multicolumn{1}{c}{Released} & & 29.45 & \textbf{10.01} & 46.25 & 57.86 \\
        \cmidrule(lr){2-7}
        & \multirow{2}{*}{~~Generated~~} & DreamBooth & 49.19 & 41.93 & 37.57 & \textbf{37.21} \\
        &                     & Textual Inversion & 92.87 & \textbf{62.67} & 102.32 &  79.58\\
        \bottomrule
        \end{tabular}
    }
\end{table}

\subsection{Influence on images quality}
\label{exp:quality}
In this subsection, we assess how FT-Shield affects the quality of both protected and generated images. Given that the impact of the watermark on image quality is irrelevant to the watermark detection setting, we conduct a collective comparison of FT-Shield with the two baseline methods.
We demonstrate the average of the FID metric for each transfer task across different datasets in Table~\ref{main:fid}. According to the results in Table~\ref{main:fid}, FT-Shield consistently achieves the lowest FID values in the released dataset. For the generated data, in most cases it also leads to a lighter influence on image quality. Although, in some cases, the FID of images generated by DreamBooth and Textual Inversion is relatively higher, as discussed in Section~\ref{exp:main} and~\ref{sec:moe}, FT-Shield consistently achieves higher watermark detection accuracy. 
This guarantees successful detection of unauthorized usage.

To offer a visual perspective, we also provide examples of the watermarked released images and generated images in Figure~\ref{fig:examples}. More visualizations can be found in Appendix~\ref{append:vis}. These visualizations confirm that FT-Shield's watermark is nearly imperceptible, maintaining the aesthetic integrity of both protected and generated images across various fine-tuning models. 

\subsection{Performance under insufficient fine-tuning steps}

\label{sec:steps}
\begin{table}[tb]
    \centering
    \captionsetup{font=small}
    \caption{\small Detection Rate (TPR) under fewer fine-tuning steps}
    \resizebox{0.62\textwidth}{!}{
        \begin{tabular}{c|c|ccc}
            \toprule
            ~steps~& ~FID~ & Ours (4/255) & \makecell{Gen-Watermark\\ \citep{ma2023generative}} & \makecell{DIAGNOSIS\\~\citep{wang2024diagnosis}}\\ 
            \midrule
            10   & 86.05 & 56.17\% & 32.67\% & 3.50\% \\ 
            20 & 83.90  & 66.00\% & 33.50\% & 7.33\% \\ 
            50 & 67.32 & 65.96\% & 52.67\% & 2.00\% \\
            100 & 59.42  & 66.94\% & 41.00\% & 1.83\% \\
            200&  49.84 & 76.50\%  & 57.97\% & 13.33\% \\
            300 & 45.93 & 97.17\% & 66.67\% & 54.67\% \\
            500 & 34.66 & 98.17\% & 85.67\% & 79.55\% \\
            800 & 35.25 & 100.00\%   & 99.72\% & 93.83\% \\
            \bottomrule
        \end{tabular}
        \label{exp:steps}
    }

\end{table}
In this subsection, we provide more evidence that the watermarks of FT-Shield can be better assimilated by the diffusion models at the early stage of the fine-tuning process. Based on DreamBooth, we conduct model fine-tuning with fewer steps compared with standard experiments. Then we apply the watermark detector tailored to DreamBooth to calculate the detection rate (TPR) of the watermark. We also apply FID to evaluate the extent to which the model learns the target style in different steps. The FIDs are derived by comparing images generated at each step against those produced at the completion of fine-tuning. The experiments are mainly done with the style transfer tasks using the paintings of artist Claude Monet. The results are demonstrated in Table~\ref{exp:steps}. 

As shown in Table~\ref{exp:steps}, FT-Shield consistently achieves the highest detection rate when the fine-tuning steps are insufficient. Even when the fine-tuning steps are as few as 10, the detection rate can achieve 56\%. With only 300 steps, the TPR can achieve nearly 100\%. In comparison, the two baseline methods require many more steps to achieve a high detection rate. By observing the change of FID, it can be found that the style has been fully assimilated by the model at the step of 500. At this moment, the TPR of FT-shield is close to 100\%, while the other two methods only achieve TPR to be 85.67\% and 79.55\%. This comparison indicates that our FT-Shield provides more robust copyright protection for data.

\subsection{Robustness of watermark}

\label{sec:robust}
The robustness of a watermark refers to its ability to remain recognizable after undergoing various modifications, distortions, or attacks. It is a crucial property of watermark because during the images' circulation, the watermarks may be distorted by some disturbances, such as JPEG compression. The data offender may also use some methods to remove the watermark. In this subsection, we show that our watermark can be robust against multiple types of corruption when proper augmentation is considered in the training of the watermark detector. In the experiment in Table~\ref{exp:robust}, we consider four types of image corruptions including JPEG compression, Gaussian Noise, Gaussian Blur and Random Crop. To make our watermark robust to those corruptions, we consider using all of these four corruptions as an augmentation in the training of each watermark detector. In Table~\ref{exp:robust}, we show the accuracy of FT-Shield-Specific
(average of True Positive Rate and True Negative Rate) which are trained with or without augmentation on the corrupted images. The performance of the watermark detector on the corrupted images is substantially improved after the augmentation is applied during the training of the detector. After the augmentation, the classifier can achieve performance near 100\% against all the corruptions in DreamBooth's images. Even in the images generated by LoRA, where the classifier performs the worst, the accuracy can still be consistently higher than 84\%.
\begin{table}[t]
\centering

\captionsetup{font=small}
\caption{\small Robustness of the watermark against different image corruptions}

\label{exp:robust}
\resizebox{0.85\textwidth}{!}{
    \begin{tabular}{c|cc|cc|cc|cc}
        \toprule
         Corruption & \multicolumn{2}{c|}{DreamBooth} & \multicolumn{2}{c|}{Textual Inversion} & \multicolumn{2}{c|}{Text-to-image} & \multicolumn{2}{c}{LoRA} \\
        \cmidrule{2-9}
       Type & w/o aug. & w/ aug. & w/o aug. & w/ aug. & w/o aug. & w/ aug. &  w/o aug. & w/ aug. \\ 
        \midrule
JPEG Comp.   & 63.83\% & 99.00\% & 86.42\% & 96.58\%& 61.08\% &93.67\% & 79.42\% & 91.08\% \\ 
Gaussian Noise & 68.50\% &99.25\% & 90.17\% & 97.75\%& 91.08\% & 91.67\% & 75.83\% & 86.17\% \\
Gaussian Blur & 45.17\% & 99.25\% & 75.58\% & 97.67\% & 92.92\% & 95.42\% & 93.08\% & 91.08\%\\ 
Random Crop &83.83\% & 99.08\% &86.50\%& 96.50\% & 73.25\% & 88.00\% & 71.58\% & 84.67\% \\
\bottomrule

\end{tabular}}

\end{table}


\section{Conclusion}

In this paper, we proposed a novel watermarking method to safeguard images' IP against the fine-tuning of text-to-image diffusion models. To ensure that the watermark can be efficiently and accurately assimilated by the diffusion model, we proposed an algorithm for watermark generation which incorporates the fine-tuning loss of diffusion models in the training loss of watermark. Meanwhile, we introduce a MoE strategy for the watermark detection to enhance its adaptability to diverse fine-tuning methods. Empirical results demonstrates the effectiveness of our method and its superiority over the existing watermarking methods. 

\bibliographystyle{paper}
\bibliography{reference}

\newpage
\appendix
\section{Algorithm}
\label{append:alg}
The detailed algorithm for the training of the watermark (Equation~\ref{eqn:obj}) is demonstrated as below.
\begin{algorithm}
    \renewcommand{\algorithmicrequire}{\textbf{Input:}}
    \renewcommand{\algorithmicensure}{\textbf{Output:}}
    \caption{Optimization for watermark $\delta_i$}
    \label{alg}
    \begin{algorithmic}[1]
		\Require Protected dataset $\{x_i\}_{i\in[n]},$ Captions for protected dataset $\{c_i\}_{i\in[n]},$ Initialized watermark $\{\delta_{i,0}\}_{i\in[n]},$ Pretrained text-to-image diffusion model with parameters $\theta_1, \theta_2$ ($\theta_1$ denotes the unet part and $\theta_2$ denotes the other parts), watermark budget $\eta$, diffusion model learning rate $r$, batch size $bs$, PGD step $\alpha$ and epoch $E$\
        \Ensure Optimal watermark $\{\delta_{i}^*\}_{i\in[n]}$
        \For {Epoch=1 to E}
            \For {Batch from $\{x_i\}_{i\in[n]}$} 
            \State $ \theta_1^* \gets \theta_1$
            \For {1 to 5}
            \State $ \theta_1^* \gets \theta_1^*-r\frac{\partial}{\partial \theta_1^*} {L}_{d m}\left(\theta_1^*, \theta_2, x_{1:bs}, c_{1:bs}\right)$ \textcolor{lightgray}{~~~~~~~//~Use clean images to update the unet}
            \EndFor
            \For {1 to 5}
            \State $ \delta_{1:bs} \gets  \delta_{1:bs}-\alpha sign\{ \frac{\partial}{\partial \delta_{1:bs}} {L}_{d m}\left(\theta_1^*, \theta_2, x_{1:bs}+\delta_{1:bs}, c_{1:bs}\right)\}$
            \State $ \delta_{1:bs} \gets Proj_{\|\delta_{1:bs}\|_\infty\leq \eta}(\delta_{1:bs}) $ \textcolor{lightgray}{~~~~~~~//~PGD to update watermark}
            \EndFor
            \For {1 to 5}
            \State $ \theta_1 \gets \theta_1- r \frac{\partial}{\partial \delta_{1:bs}} {L}_{d m}\left(\theta_1, \theta_2, x_{1:bs}+\delta_{1:bs}, c_{1:bs}\right)$ \textcolor{lightgray}{//~Use watermarked images to update the unet}
            \EndFor
            \EndFor
        \EndFor
    \end{algorithmic}
\end{algorithm}

\section{Additional Details about the Fine-tuning Methods}
\label{append:exp}
In the experiments of this paper, we considered four Fine-tuning methods of text-to-image models including DreamBooth, Textual Inversion, Text-to-Image and Text-to-Image-LoRA for style transfer and object transfer tasks. More details about the setting of these fine-tuning methods are provided as below.

\begin{itemize}[leftmargin=*]
\item \textbf{DreamBooth ~\cite{ruiz2023dreambooth}}: DreamBooth is a fine-tuning method to personalize text-to-image diffusion models. It mainly focus on fine-tuning the unet of the diffusion models' structure with a prior preservation loss to avoid overfitting and language-drift.
Whether to update the text-encoder within in the text-to-image model structure is an open option. In the experiment in this paper, we update both the unet and the text-encoder with learning rate to be 2e-6, batch size to be 1 and maximum fine-tuning stpes to be 800. For style transfer task, we use ``[V]" as the unique identifier for the specific style and incorporate ``[V]" in the prompts  in the sampling process to instruct the model to generated images following this style. Similarly, for object transfer, we use "sks" as the identifier.

\item \textbf{Textual Inversion~\cite{gal2022image}}: Textual Inversion is another text-to-image diffusion model personalization method. It focuses on adding a new token which is connected to a specific style or object to the vocabulary of the text-to-image models. This work by using a few representative images to fine-tune the text embedding of the pipeline's text-encoder. In our experiment, we set the fine-tuning learning rate to be 5.0e-04, batch size to be 1 and maximum fine-tuning steps to be 1500. We use ``[V]" as a placeholder to represent the new concepts that the fine-tuning process learn and also incorporate it in the prompts in the sampling process.

\item \textbf{Text-to-Image}: Text-to-Image Fine-Tuning Method is a simple implementation of the fine-tuning of text-to-image diffusion models. It fine-tunes the whole unet structure with a dataset which contains both the images and the captions describing the contents of the images. An identifier ``[V]" is also required in the captions in the fine-tuning and sampling procedure.
In our experiment, we set the learning rate for fine-tuning  to be 5e-06, the batchsize to be 6 and the maximum fine-tuning steps to be 300.

\item \textbf{LoRA~\cite{hu2021lora}}: LoRA works by adding pairs of rank-decomposition matrices to existing weights of the UNet and only train the newly added weights in the fine-tuning process. It also required a image-caption pair dataset for fine-tuning and and the identifier ``[V]" in the cations. In experiments, we set the learning rate to 5e-06, batch size to 6 and maximum training steps to 3000.

\end{itemize}

\section{Additional Visualization of Watermarks}
\label{append:vis}

\begin{figure}
    \centering
    \includegraphics[width=1\textwidth]{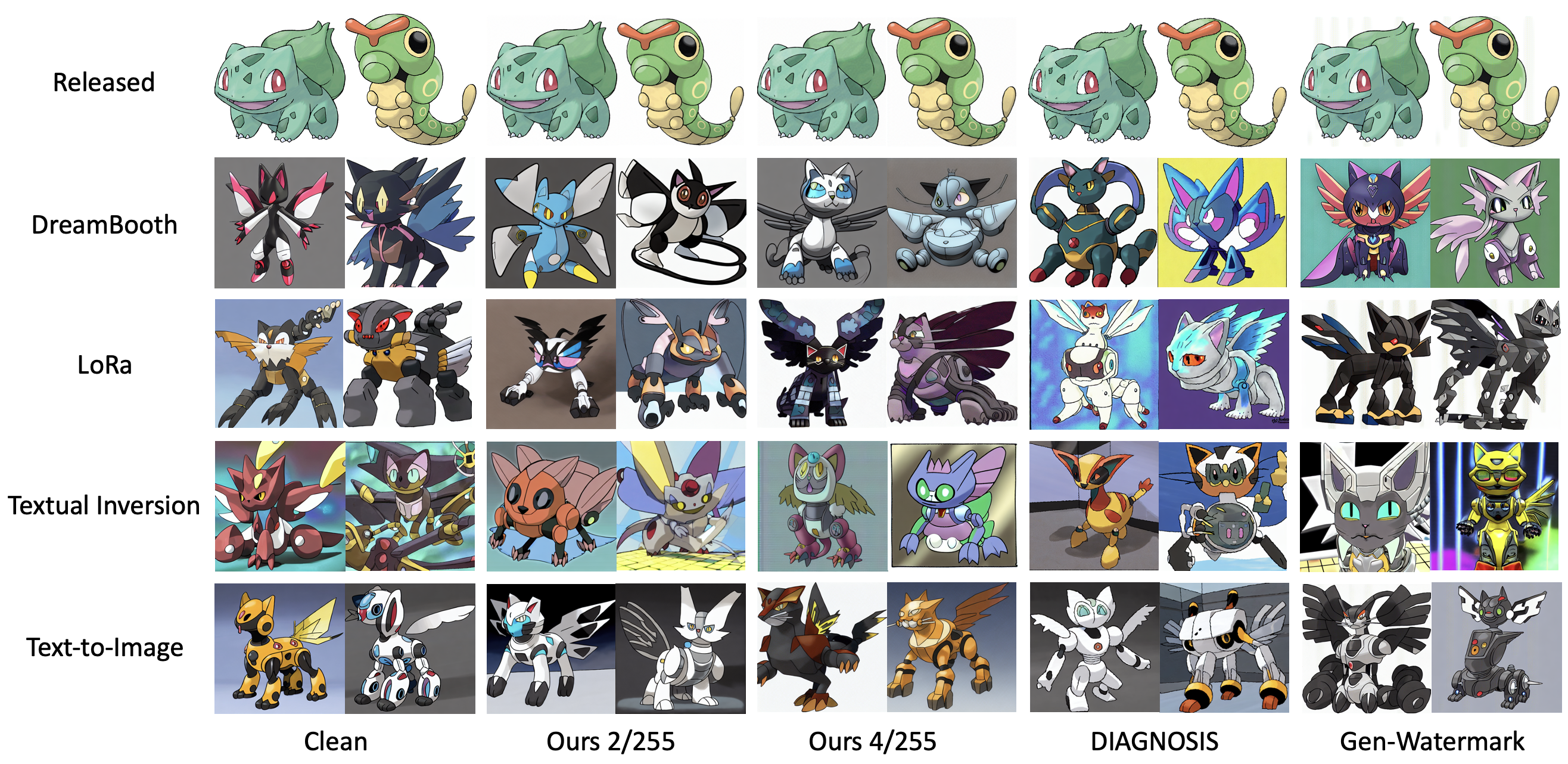}
    \caption{{Examples of watermarked images (first line) and  images generated through domain adaptation for Pokemon imagery (other lines). The prompt of generation: A robotic cat with wings.}}
    \label{fig:vis_poke}
\end{figure}

\begin{figure}
    \centering
    \includegraphics[width=1\textwidth]{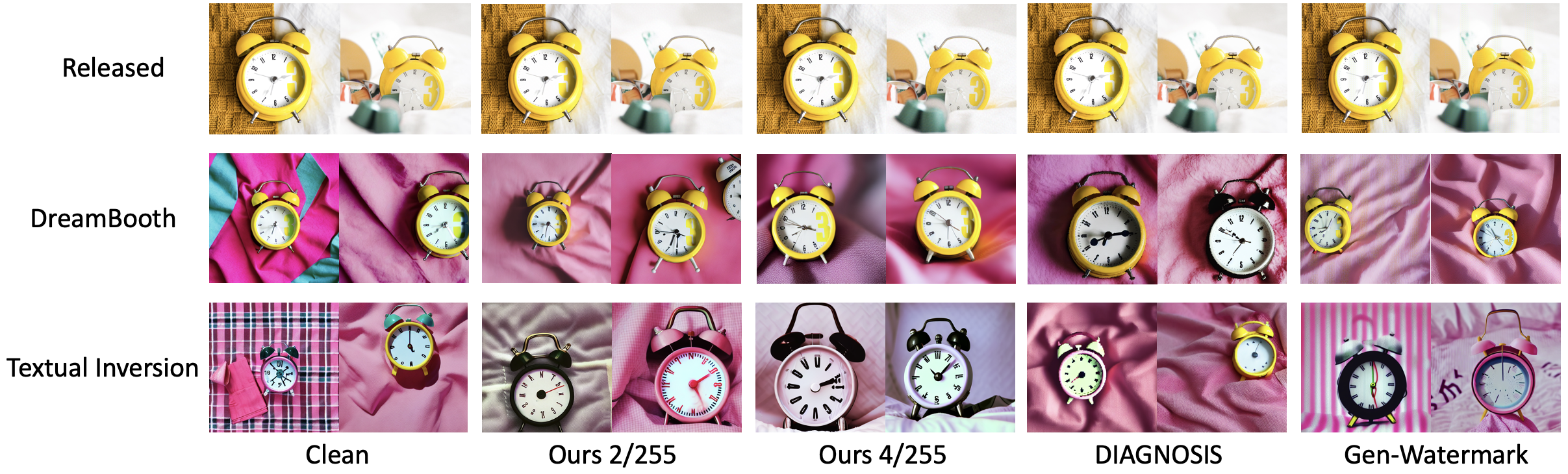}
    \caption{{Examples of watermarked images (top row) and images generated from object transfer (clock) shown in subsequent rows. The prompt of generation: A sks clock on top of pink fabric.}}
    \label{fig:vis_clock}
\end{figure}

\begin{figure}
    \centering
    \includegraphics[width=1\textwidth]{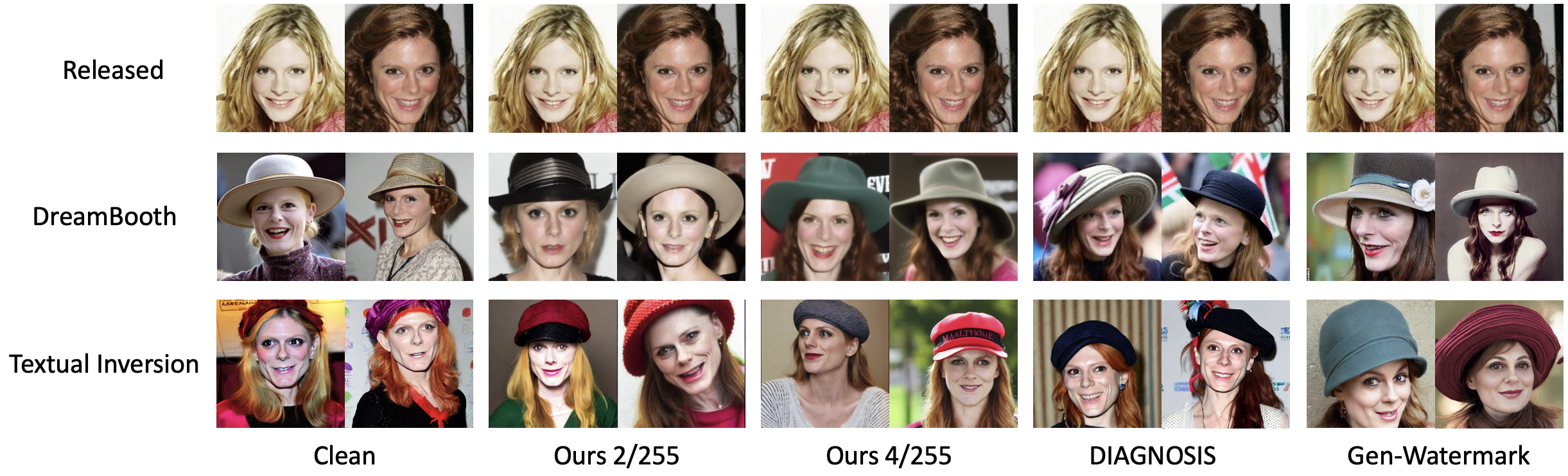}
    \caption{{Examples of watermarked images (first line) and images generated from face transfer (other lines). The prompt of generation: A photo of sks person wearing a vintage hat.}}
    \label{fig:vis_face}
\end{figure}

In this section, we provide some additional visualization of watermarks as in Figure~\ref{fig:vis_poke}, \ref{fig:vis_clock}, and \ref{fig:vis_face}.


\section{Ablation Studies and Additional Empirical Results}

\label{sec:abl}

\textbf{Reduced watermark rate.}
Watermark rate refers to the percentage of a dataset that is protected by watermarks. In real practice, the data protector may have already released their unmarked images before the development of the watermark technique. Therefore, it is necessary to consider the situation where the watermark rate is not 100\%.
In this subsection, we demonstrate the effectiveness of FT-Shield when the watermark rate is lower than 100\%. The experiments are mainly based on the style transfer task using the paintings by artist Louise Abbema. The results are presented in Table~\ref{exp:rate}.
As shown in the table, as the proportion of the watermarked images in the training set decreases, the TPR also decreases. This is within expectation because when there are fewer watermarked images in the protected dataset, it is harder for the watermark to be assimilated by the diffusion model. Nonetheless, our method consistently achieves better performance than baselines.
With a watermark rate of 80\%, it achieves a detection rate close to 100\% across all fine-tuning methods. Even when the watermark rate is reduced to 20\%, FT-Shield still maintains detection rates higher than 80\% across all the fine-tuning methods except LoRA. Although the watermark detection rate for LoRA’s generated images experienced the most substantial decline, it remains much higher than the two baseline methods.

\begin{table}[tb]
    \centering
    \captionsetup{font=small}
    \caption{\small Watermark detection rate (TPR) under different watermark rates (`FS' and `FM' refer to `FT-Shield-Specific' and `FT-Shield-MoE', respectively, `GW' stands for `Gen-Watermark'~\cite{ma2023generative}, and `DN' represents `DIAGNOSIS'~\cite{wang2024diagnosis}).}

    \label{exp:rate}
    \resizebox{1.02\textwidth}{!}{
    \begin{tabular}{c|cc|cc|cc|cc|cc|cc|cc|cc}
        \toprule
        \multirow{2}{*}{\makecell{WM \\ Rate}}  & \multicolumn{4}{c|}{DreamBooth} & \multicolumn{4}{c|}{Textual Inversion} & \multicolumn{4}{c|}{Text-to-image} & \multicolumn{4}{c}{LoRA} \\
       \cmidrule{2-17}
        & FS & GW &  FM & DN & FS & GW &  FM& DN &FS & GW & FM& DN & FS & GW & FM & DN\\
        \midrule
        100 \% &  99.66\% & 98.47\%&  99.67\%&91.50\%& 96.54\% & 87.78\%&96.83\%&88.28\%&98.74\% & 93.33\% & 99.33\% &88.89\%&97.49\% & 85.83\% & 98.33\%&87.83\%\\
        80 \% & 99.58\%  & 97.22\%&98.83\% & 90.68\%&97.75\% &84.86\%&95.00\%&88.35\% &96.92\%& 80.70\% & 96.00\%  & 63.73\%&95.00\%&  78.34\%& 91.67\%&75.21\%\\
        50 \%  &  95.92\% & 94.45\% & 87.33\%&54.91\%&92.92\% &83.20\% & 88.50\%& 37.60\%&86.25\% &  77.50\%& 83.67\%&48.75\%&73.75\% &  66.11\%& 66.17\%&64.39\%\\
        20 \%  & 86.42\%  &92.78\% & 83.50\%&13.81\%&88.25\% &81.25\% & 85.83\%&19.30\% &83.33\%& 74.45\% & 81.00\%&32.11\%&63.83\% & 54.45\% & 56.50\%&44.76\%\\
        
        \bottomrule

    \end{tabular} }

\end{table}

\label{appd:ablation}
\begin{table}[tb]
    \centering

    \captionsetup{font=small}
    \caption{\small Performance of watermark detector trained without augmentation}

    \resizebox{0.4\textwidth}{!}{
        \begin{tabular}{llcc}
            \toprule
             & & \makecell{Ours \\ ($4/255$)} & \makecell{Ours \\ ($2/255$)} \\
            \midrule
            \multirow{2}{*}{DreamBooth}        & TPR$\uparrow$   & 84.67\% & 79.06\% \\
            & FPR$\downarrow$ & 9.56\% & 9.61\% \\
            \midrule
            \multirow{2}{*}{Textual Inversion} & TPR$\uparrow$   & 54.83\% & 47.61\% \\
            & FPR$\downarrow$ & 3.83\% & 13.22\% \\
            \midrule
            \multirow{2}{*}{Text-to-image}     & TPR$\uparrow$   & 37.89\% & 46.00\% \\
            & FPR$\downarrow$ & 2.89\% & 15.50\% \\
            \midrule
            \multirow{2}{*}{LoRA}              & TPR$\uparrow$   & 44.06\% & 53.61\% \\
            & FPR$\downarrow$ & 4.72\% & 19.67\% \\
            \bottomrule
        \end{tabular}
    }
    \label{exp:noaug}
\end{table}

\textbf{Detector trained without data augmentation.}
As discussed in Section~\ref{detector}, it is necessary to use the generated data from the fine-tuned diffusion model to augment the dataset used for the watermark detector's training. Table~\ref{exp:noaug} demonstrates the performance of the classifier if there is no augmentation (based on the style transfer task). The classifier is simply trained on the dataset which contains the clean and watermarked protected images. According to the results demonstrated in Table~\ref{exp:noaug}, when there is no augmentation, the watermark detector can successfully detect some watermarked images on the generated set, especially those generated by DreamBooth. However, the performance will be much worse than the ones with augmented data. This difference demonstrates the necessity to conduct augmentation when training the watermark detector.

\textbf{Transferability of tailored watermark detector.} We provide some additional results (results for the object transfer tasks) regarding the transferability of tailored watermark detector here. The results are shown in Table~\ref{exp:trans}, indicating that the performance of the detectors is greatly reduced when the classifier is applied to the images generated by a different fine-tuning method. 

\textbf{Robustness of watermark (FT-Shield-MoE).} 
Given the similar performance of FT-Shield-MoE to FT-Shield-Specific, in Section~\ref{sec:robust} we primarily 
present the detection performance of FT-Shield-Specific. Nonetheless, detection performance of FT-Shield-MoE is also provided, as seen in Table~\ref{exp:robust_moe}. Comparing Table~\ref{exp:robust} and~\ref{exp:robust_moe} we can see that, in general, the detection performance of FT-Shield-MoE is similar to that of FT-Shield-Specific. Both FT-Shield-Specific and FT-Shield-MoE achieve good performance in watermark detection on corrupted generated images.

\begin{table}[tb]
    \centering
    \captionsetup{font=small}
    \caption{\small Transferability of the watermark detectors in object transfer}

    \resizebox{0.6\textwidth}{!}{
        \begin{tabular}{l|l|cc}
            \toprule
          budget & & DreamBooth & Textual Inversion \\
            \midrule
            \multirow{2}{*}{2/255} & DreamBooth        & 98.24\%&57.28\%  \\
            &Textual Inversion & 68.23\% & 97.63\%\\
           \midrule
            \multirow{2}{*}{4/255} & DreamBooth        & 98.85\% &73.77\%  \\
            &Textual Inversion & 72.05\% & 98.03\% \\
            \bottomrule
        \end{tabular}
        \label{exp:trans}
    }

\end{table}

\begin{table}[t]
\centering

\captionsetup{font=small}
\caption{\small Robustness of the watermark against different image corruptions (detected by FT-Shield-MoE)}

\label{exp:robust_moe}
\resizebox{0.85\textwidth}{!}{
    \begin{tabular}{c|cc|cc|cc|cc}
        \toprule
         Corruption & \multicolumn{2}{c|}{DreamBooth} & \multicolumn{2}{c|}{Textual Inversion} & \multicolumn{2}{c|}{Text-to-image} & \multicolumn{2}{c}{LoRA} \\
        \cmidrule{2-9}
       Type & w/o aug. & w/ aug. & w/o aug. & w/ aug. & w/o aug. & w/ aug. &  w/o aug. & w/ aug. \\ 
        \midrule
JPEG Comp.   & 62.17\% & 98.92\% & 75.42\% & 98.33\%& 85.33\% &94.33\% & 76.17\% & 90.08\% \\ 
Gaussian Noise & 84.58\% &98.67\% & 95.00\% & 96.67\%& 71.33\% & 88.42\% & 78.33\% & 85.42\% \\
Gaussian Blur & 89.75\% & 99.17\% & 69.58\% & 95.42\% & 92.17\% & 94.00\% & 89.03\% & 91.50\%\\ 
Random Crop &82.83\% & 98.83\% &73.75\%& 95.83\% & 71.92\% & 84.08\% & 73.67\% & 86.92\% \\
\bottomrule

\end{tabular}}

\end{table}

\section{Prompts Used for Images Generation}
\label{appd:prompts}

In the following, we provide the prompts used in image generation in our experiments.
\subsection{Prompts Used for Face Object Transfer}
\begin{adjustwidth}{0cm}{-1.1cm}

\fontsize{7.6pt}{7.6pt}\selectfont
\begin{tabular}{p{0.38\linewidth}p{0.55\linewidth}}
    \toprule
    \midrule
    A photo of [V] laughing heartily & A photo of [V] by the beach at sunset \\
    A photo of [V] wearing a vintage hat & A photo of [V] with a glass of wine \\
    A photo of [V] reading a thick book & A photo of [V] wearing a graduation cap \\
   A photo of [V] with ear rings  & A photo of [V] holding a vintage camera \\
    A photo of [V] holding a coffee cup & A photo of [V] holding a bouquet of flowers\\
    A photo of [V] in a classroom & A photo of [V] wearing A winter scarf and gloves \\
    A photo of [V] on a boat   & A photo of [V] wearing oversized sunglasses \\
    A photo of [V] in the jungle & A photo of [V] in front of a flower field  \\
    A photo of [V] with a kitten & A photo of [V] amidst colorful autumn leaves \\
    A photo of [V] in a sunny park &A photo of [V] holding a bottle of water \\
    A photo of [V] in her bedroom &  A photo of [V] surrounded by festive balloons \\
    A photo of [V] with upset face  &  A photo of [V] with a colorful parrot on the shoulder\\
    A photo of [V] with blunt-cut bangs & A photo of [V] with straight black hair \\
    A photo of [V] in the street & A photo of [V] in front of a window \\
    A photo of [V] with short hair & \\
    A photo of [V] in a library, surrounded & by towering bookshelves \\
    \bottomrule
\end{tabular}

\end{adjustwidth}

\subsection{Prompts Used for Lifeless Object Transfer}
\begin{adjustwidth}{0cm}{-1.1cm}



\fontsize{8pt}{8pt}\selectfont
\begin{tabular}{p{0.4\linewidth}p{0.55\linewidth}}
    \toprule
    \midrule
    A [V] in the snow & A [V] with a wheat field in the background \\
    A [V] on the beach & A [V] with a tree and autumn leaves in the background \\
    A [V] on a cobblestone street & A [V] with the Eiffel Tower in the background \\
    A [V] on top of pink fabric & A [V] on top of green grass with sunflowers around it \\
    A [V] on top of a wooden floor & A [V] on top of a mirror \\
    A [V] with a city in the background & A [V] on top of a dirt road \\
    A man with a [V] & A [V] on top of a white rug \\
     A red [V] & A [V] with a blue house in the background  \\
    A cube shaped [V] & A [V] placed beside a window \\
    A girl holding a [V] & A [V] on a desk \\
    A [V] on a chair & A [V] beside a computer \\
    A [V] on the top of a roof & A [V] in a box \\
    A [V] on a bed & A [V] with a mountain in the background \\
    A [V] on a desk & A [V] on a cliff overlooking the sea \\
    A [V] placed on a bookshelf & A [V] under a tree \\
    \bottomrule
\end{tabular}

\end{adjustwidth}

\subsection{Prompts Used for Style Transfer (Arts)}
\begin{adjustwidth}{0cm}{-1.1cm}

\fontsize{8pt}{8pt}\selectfont
\begin{tabular}{p{0.42\linewidth}p{0.6\linewidth}}
    \toprule
    \midrule
    A lady reading on grass in the style of [V] & Anglers on the Seine River in the style of [V] \\
    Flower field in the style of [V] & Haystacks in winter mornings in the style of [V] \\
    Iris in the style of [V] & Mother and her child in a garden in the style of [V] \\
    Red Boats at Argenteuil in the style of [V] & Saint Lazar Railway Station in the style of [V] \\
    Sunflowers in the style of [V] & A man in a suit with a beard in the style of [V]\\
    Waterfall in the style of [V] & Boats at rest at petit gennevilliers in the style of [V] \\
    Eese in the creek in the style of [V] & Claude haystack at giverny in the style of [V] \\
   Meadow with poplars in the style of [V]   & Olive tree wood in the moreno garden in the style of [V] \\
    A fountain in the style of [V] & A bottle of champagne in an ice bucket in the style of [V] \\
    Snow scene in the style of [V] & An Italian vineyard at midday in the style of [V] \\
    The artist house in the style of [V] &  Cherry blossoms in full bloom in the style of [V]  \\
      The cabin in the style of [V] & The bodmer oak fontainebleau in the style of [V]\\
    Sunrise in the style of [V] & The sea at saint adresse in the style of [V] \\
    The seine in the style of [V] &Birds taking flight from a tree in the style of [V] \\
    Walk in the meadows in the style of [V]& The summer poppy field in the style of [V]  \\
Rough sea in the style of [V]    & An old man playing the violin in the style of [V] \\
    Woman in a garden in the style of [V] & Swans gliding on a serene pond in the style of [V] \\
    A growling tiger in the style of [V] & Majestic castle overlooking a valley in the style of [V] \\
    Waterloo bridge in the style of [V]  & Wild horses galloping on the shore in the style of [V] \\
  The boat studio in the style of [V]  & Bustling train station in the 1900s in the style of [V] \\
    The sheltered path in the style of [V]   & Portrait of a Woman with Low Neckline in the style of [V]  \\
    Ballerinas rehearsing in the style of [V] & Market day in a provincial town in the style of [V] \\
     A goat on grass in the style of [V] & The old lighthouse by the cliff in the style of [V]  \\
    Two butterflys in the style of [V] &  Moonlit night over a calm sea in the style of [V] \\
    A frog on a lotus Leaf in the style of [V] & A camel team in the desert in the style of [V] \\
    An eggplant on vines in the style of [V] & Windmills on the Flower Field in the style of [V]  \\
    Alm tree at bordighera in the style of [V] &  A dog waiting for the owner in the style of [V] \\
   A view of mountain in the style of [V]  & Two pandas eating bamboos in the style of [V] \\
    Grapes on a vine in the style of [V] &  A woman admiring lotus flowers in the style of [V]\\
 The Cliffs of Etelta in the style of [V]     & The side face of a red-haired woman in the style of [V] \\

    \bottomrule
\end{tabular}

\end{adjustwidth}

\small \textbf{Remarks}: In the sampling procedure of DreamBooth and Textual Inversion, the prompts need to be added with the word "painting", e.g., A lady reading on grass in the style of [V] painting. 

\subsection{Prompts Used for Style Transfer (Pokemon)}
\begin{adjustwidth}{0cm}{-1.1cm}
\fontsize{8pt}{8pt}\selectfont
\begin{tabular}{p{0.42\linewidth}p{0.6\linewidth}}
    \toprule
    \midrule
    A robotic cat with wings  & A blue and white dinosaur with wings  \\
A phoenix with icy feathers & A cartoon sunflower with a happy face  \\
A steel-spined hedgehog & A deer with colorful feathers on it's head  \\
A small, furry creature with large eyes & A gray cartoon character with a black tail \\
A fire-breathing fox & A pokemon ball with a butterfly on top of it  \\
A jellyfish-like creature & A red and white toy with a blinking green eye \\
A robotic unicorn with a laser horn & A small, insect-like creature with petal-like wings \\
A robotic dog with butterfly wings &A magical gnome that walks at night \\
A tree-like creature with glowing eyes & A bioluminescent jellyfish with a galaxy pattern inside\\
A creature made of clouds &  A cartoon character with claws \\
A creature with butterfly-like wings & A yellow and orange pokemon with big red eyes \\
A fairy-tale dragon &A miniature dragon made of living crystal \\
An armored turtle with a spiked tail &A will-o'-the-wisp in the form of a playful kitten \\
A cyborg rabbit with blue eyes & A floating island turtle with a mini ecosystem on its back \\
A robotic owl with holographic wings & A dog with a wheel in his hand   \\
A cyborg penguin with jet engines & A mechanical elephant with hover discs for feet \\
A robot bear with solar panels as fur &A spectral wolf with aurora-like fur \\
A blue and yellow insect &An origami crane that comes to life \\
A pink dog with red ears sitting down & A chameleon with digital camouflage \\
A goat with a bell on its head & A peacock with holographic tail feathers \\
A red and white dragonfly & A saber-toothed cat with plasma claws \\
A blue spider with a green cap & A clockwork bird with wings of stained glass \\
A green and yellow bear & A starry narwhal with a radiant horn \\
A green and red dragon & An ice-cream shaped panda with flavors as fur colors \\
A red and orange pokemon &A shadow fox that morphs into smoke  \\
A cute animal with horns & A wind-up mouse with antique clockwork gears \\
A colorful butterfly &A coral reef mermaid with sea glass and shells \\
A pair of owls with orange wings &  A robot with a green body, yellow arms, and a red head  \\
A bamboo dragon that grows real leaves &\\
A glowing, ethereal deer with constellation & patterns in its fur \\
A neon-furred squirrel with jetpack wings, & channeling the futuristic vibe\\ 
 An aquatic creature with the features of a & shark and a dolphin \\
    \bottomrule
\end{tabular}
\end{adjustwidth}


\end{document}